\documentclass[runningheads]{llncs}
\usepackage{graphicx}
\usepackage{listings}

\usepackage{wrapfig}
\begin{document}
%

\title{Embedding Vector Differences \\ Can Be Aligned With \\ Uncertain Intensional Logic Differences}

%

\author{Ben Goertzel, Mike Duncan, Debbie Duong, Nil Geisweiller, Hedra Seid, Abdulrahman Semrie, Man Hin Leung, Matthew Ikle'}
\authorrunning{Goertzel et al.}
%
 \institute{SingularityNET Foundation }
%
\maketitle              
\begin{abstract}
The DeepWalk algorithm is used to assign embedding vectors to nodes in the Atomspace weighted, labeled hypergraph that is used to represent knowledge in the OpenCog AGI system, in the context of an application to probabilistic inference regarding the causes of longevity based on data from biological ontologies and genomic analyses.    It is shown that vector difference operations between embedding vectors are, in appropriate conditions, approximately alignable with  ``intensional difference'' operations between the hypergraph nodes corresponding to the embedding vectors.   This relationship hints at a broader functorial mapping between uncertain intensional logic and vector arithmetic, and opens the door for using embedding vector algebra to guide intensional inference control. 
\end{abstract}
%
%
%
\section{Introduction}

Graph embedding algorithms assign vectors to nodes of a graph, with elegant properties such as: Nodes which are similar according to the graph topology and geometry get similar embedding vectors.  

Word embedding vectors derived from natural language corpora via algorithms like word2vec display desirable "vector arithmetic" properties (e.g. man-woman = king-queen, where by "man" in the equation is meant the embedding vector for the word "man").   

An interesting question is then: If we have a natural notion of "semantic difference" between nodes in a graph, do the relationships between embedding vector differences reflect corresponding relationships between node semantic differences?

We present preliminary proof of concept results suggesting that, if embedding is done appropriately, sometimes the answer is yes.  

\section{Explorations with the Bio-Atomspace}

We have conducted this investigation in the context of our utilization of the OpenCog Atomspace \cite{EGI1} \cite{EGI2} -- a weighted, labeled hypergraph AI knowledge store -- to conduct probabilistic logical inference regarding the genomics of longevity.   We use the "Bio-Atomspace"  -- an Atomspace filled with knowledge from multiple bio-ontologies, and with results from statistical and machine learning analysis of various genomics datasets from longevity-related studies (see \cite{PLNBio} for prior work with an earlier version of Bio-Atomspace).  Application of the Probabilistic Logic Networks (PLN) engine \cite{PLN} to the Bio-Atomspace produces uncertain logical explanations for the connections found by machine learning algorithms between certain gene variations or expressions and combinations thereof and phenotypes such as longevity.   

A very simple example of the inferences PLN conducts in this context is as follows:

\begin{verbatim}
;; Inference trail of
;;
;; (MemberLink (stv 0.12426852 0.061859411)
;;    (GeneNode "ITPR3")
;;    (ConceptNode "HAGR increased expression-with-aging GeneSet")
;; )
?
(ListLink
   (ListLink
      (DefinedSchemaNode "intensional-similarity-direct-introduction-rule")
      (ConceptNode "GO:0050794" (stv 0.55316436 0.96080161))
      (NumberNode "1")
   )
   (ListLink
      (DefinedSchemaNode "intensional-similarity-to-member-rule")
      (IntensionalSimilarityLink (stv 0.092158662 0.67346939)
         (ConceptNode "GO:0030889" (stv 0.00081595186 0.96080161))
         (ConceptNode "GO:0050794" (stv 0.55316436 0.96080161))
      )
      (NumberNode "89")
   )
   (ListLink
      (DefinedSchemaNode "intensional-similarity-property-deduction-rule")
      (IntensionalSimilarityLink (stv 0.13080897 0.13469388)
         (GeneNode "FCGR2B")
         (GeneNode "ITPR3")
      )
      (NumberNode "1345")
   )
)
\end{verbatim}

\noindent -- this inference basically explains why gene ITPR3 has increased gene expression in aged individuals, via noting its possession of many similar properties to gene ITPR2 (which has increased gene expression in aged individuals); and noting that it belongs to Gene Ontology category 50794, which is similar to Gene Ontology category 30889, which is known to be related to aging.   Many more complex and subtle inferences are constructed as PLN does its work on the Bio-Atomspace, but they involve similar players.

In this bio-AI setting, one relevant measure of semantic difference is the "intensional difference" between two concept-representing hypergraph nodes, which measures the quantity of informative properties held by one of a pair of nodes but not the other. 

We present exploratory analysis showing that in some cases of real-world relevance, intensional difference between concept nodes behaves similarly to vector difference between the embedding vectors corresponding to the concept nodes.

If these preliminary observations hold up more broadly, this will be highly valuable for inference control.   It suggests that one may be able to guide intensional inference by directing a logic engine to roughly follow a vector between the embedding vector of the premises and the embedding vector of the desired conclusion.

\section{DeepWalk on Atomspace}

For producing vector embeddings from the OpenCog Atomspace, we have utilized the DeepWalk algorithm \cite{DeepWalk}  to create (e.g. 100-dimensional) numerical vectors corresponding to Atomspace nodes.  We also did some preliminary experiments with GraphCNNs, but based on our early explorations this seemed less promising so we proceeded with DeepWalk.

The rough methodology involved here is:

\begin{enumerate}
\item Paths through the Atomspace knowledge hypergraph are created and exported
\item The corpus of paths is analyzed, much as if it was a corpus of natural language sentences
\item Vectors are assigned to nodes/links based on neural-net analysis of their relationship to other nodes/links in the paths
\end{enumerate}

\noindent This allows vector-processing algorithms such as neural nets to be applied to (vectorial representations of) symbolic nodes and links, complementing and synergizing with the symbolic manipulations occurring within the Atomspace.

Two example paths from Bio-Atomspace, among the numerous fed to DeepWalk for producing its embedding vectors are:

\begin{verbatim}
['GO:0039625', 'inherits-geneontologyterm', 'GO:0044423', 
'geneontologyterm-inherited-by', 'GO:0019028', 'inherits-geneontologyterm', 
'GO:0044423', 'geneontologyterm-inherited-by', 'GO:0098025', 
'inherits-geneontologyterm', 'GO:0044423', 'inherits-geneontologyterm', 
'GO:0005575', 'has-gene-ontology-member', 'OXNAD1', 'interacts_with', 
'PROSC', 'interacts_with', 'SMS', 'interacts_with', 'BAP1']

['GO:1900826', 'has-gene-ontology-member', 'CAV3', 'is-in', 'plasma membrane',
'in-context-of', 'R-HSA-445355', 'is-context-where',
'cytoplasmic vesicle membrane',  'has', 'TRIM72', 'is-in', 
'cytoplasmic vesicle membrane', 'in-context-of',  'R-HSA-445355',
'inherits-pathway', 'R-HSA-397014', 'pathway-inherited-by',  'R-HSA-445355', 
'inherits-pathway', 'R-HSA-397014', 'pathway-inherited-by', 'R-HSA-390522', 
'inherits-pathway', 'R-HSA-397014', 'pathway-inherited-by',  'R-HSA-5576891']

\end{verbatim}

\noindent Basically, what DeepWalk does is to assign an embedding vector to a node based on which other nodes and links it occurs nearby in these various paths.   Two nodes will get similar embedding vectors if they tend to occur in similar contexts in the set of walks.  

If one imagines a sparse feature vector for each node, with each entry corresponding to the degree to which the node possesses a certain contextual feature (e.g. occurring adjacent to or shortly thereafter some other node in paths; and with a degree calculated in terms of the informativeness with which this feature allows you to distinguish the node from other nodes), then the embedding vector of a node is conceptually similar to a PCA-type embedding of this sparse feature vector.   Indeed there is evidence that PCA on these sorts of sparse feature vectors have similar behavior to word2vec type embeddings \cite{levy-goldberg-2014-linguistic}.

\subsection{Arithmetic on Atomspace Embedding Vectors}

The word2vec vector arithmetic symmetries exemplified by the case "man - woman = king - queen" mentioned above, is also observable in the Bio-Atomspace setting.

Let e.g.  $V( \textrm{B cell differentiation})$  denote the  embedding vector for the Node corresponding to the concept ``B cell differentiation"  (corresponding in this case from a Gene Ontology category of the same name).   We then find vector arithmetic identities such as

$$
V(\textrm{B cell differentiation}) - V(\textrm{T cell differentiation}) = V(\textrm{B cell proliferation}) - V(\textrm{T cell proliferation})
$$

\noindent analogous to relations found among word2vec vectors embedding natural language concepts.

The general pattern underlying these sorts of vector difference identities may be summarized as

$$
V(A \& X) - V(B \&X ) = V(A \& Y) - V(B\&Y)
$$

where e.g.

\begin{itemize}
\item A = \textrm{male}
\item B = \textrm{female}
\item X = \textrm{human}
\item Y = \textrm{top royalty}
\end{itemize}

\noindent or

\begin{itemize}
\item A = \textrm{B-cell}
\item B = \textrm{T-cell}
\item X = \textrm{differentiation}
\item Y = \textrm{proliferation}
\end{itemize}

The reason this sort of relationship might hold is conceptually quite clear.   If vector entries represent combinations of node properties, weighted by their informativeness about the node corresponding to the vector, then the identity above basically means: {\it The properties that have higher magnitude and are more informative for A than for B, retain this comparative superiority even if one restricts A and B to particular contexts like X or Y}.   I.e., the identity between differences represents an assertion that {\it the relationship between A and B is independent of X and of Y}.   Like many probabilistic independence assumptions regarding natural concepts, this will be roughly true much of the time but not all the time.

In the word2vec case, the logic of relationships between concepts like ``male'', ``female", ``human" and ``top royalty" is wholly implicit as the data involved in generating embeddings is just a sequence of sentences.   In the Bio-Atomspace case, we have explicit representations of the concepts involved and the logical relationships between them -- which we will exploit below.

It's worth emphasizing that the phenomena we study here are not peculiar to the biomedical use-case -- this is just where we happen to have initially encountered and explored them.  In fact we expect the same phenomena to occur in the domain of everyday concepts like ``male'', ``female", ``human" and so forth.   However one would need a reasonably large-scale Atomspace containing abstract, uncertain logic relations between these concepts.   We are currently engaged in research aimed at constructing such an Atomspace, one consequence of which will be to enable the same issues we explore here regarding the Bio-Atomspace to be explored in the context of everyday concepts.  

\section{Parallelism Between Intensional Difference Relationships and Embedding Vector Difference Relationships}

In OpenCog?s PLN reasoning system, we have ``intensional logic" that concerns the patterns and properties of a concept, rather than its explicit examples/members.   For instance the IntensionalInheritance between $A$ and $B$ is defined as the probabilistic (extensional) inheritance between the fuzzy set $Pat(A)$ of properties of $A$ and the fuzzy $Pat(B)$ set of properties of $B$.   The degree to which a property $p$ belongs to $Pat(A)$ is calculated as the amount of information that is given about a member of $A$ via specifying the property $p(A)$.

Along the same lines we may define

$$
IntensionalDifference(A,B) = Pat(A) - Pat(B)  
$$

\noindent (where ? denotes fuzzy set difference).   

One hypothesis we are currently exploring is that: When the vector difference identity

$$
V(A \& X) - V(B \&X ) = V(A \& Y) - V(B\&Y)
$$

\noindent approximately holds, then the intensional logic relationship

\begin{verbatim}
Similarity( {IntensionalDifference(A & X,B &X  ) , 
				IntensionalDifference(A & Y, B & Y ) )
\end{verbatim}

\noindent holds as well.

The theoretical reason here is simple: The same independence assumption that would make the vector difference identity true, would tend to make the intensional logic relationship true.

Based on evaluation of concrete examples in the Bio-Atomspace, this theoretical analysis seems to be validated.  In the example given above regarding B-cell and T-cell differentiation and proliferation, for example, we find a very high truth value for 

\begin{verbatim}
Similarity( IntensionalDifference(B-cell prolif, T-cell prolif ) , 
			IntensionalDifference(B-cell diff, T-cell diff ) )
\end{verbatim}

\noindent (where e.g. ``B-cell prolif" refers to the ConceptNode in the BioAtomspace corresponding to the Gene Ontology category named "B-cell proliferation").

\begin{figure*}[htb]
\centering
\includegraphics[width=16cm]{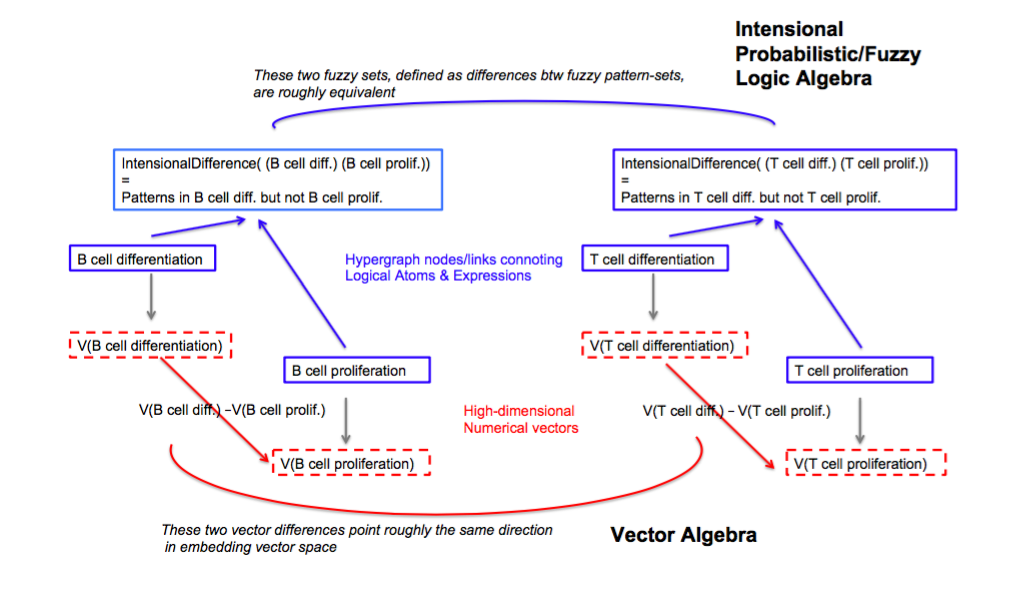}
\caption{Illustration of the alignment between relationships among vector differences and relationships among intensional logic differences}
\label{fig:morphism}
\end{figure*}

The concept of the mapping here is partially captured in Figure \ref{fig:morphism}.

This alignment may possibly be the result of a broader functorial mapping between vector algebra and uncertain intensional logic.  It is tempting to hypothesize that the DeepWalk embedding is a functor mapping the algebra of uncertain intensional logic operations (union, intersection, negation, difference) into the algebra of vector arithmetic.   The validity of this more general mapping is a subject of our current investigation.

\subsection{Potential Applications to Inference Control}

\begin{figure*}[htb]
\centering
\includegraphics[width=16cm]{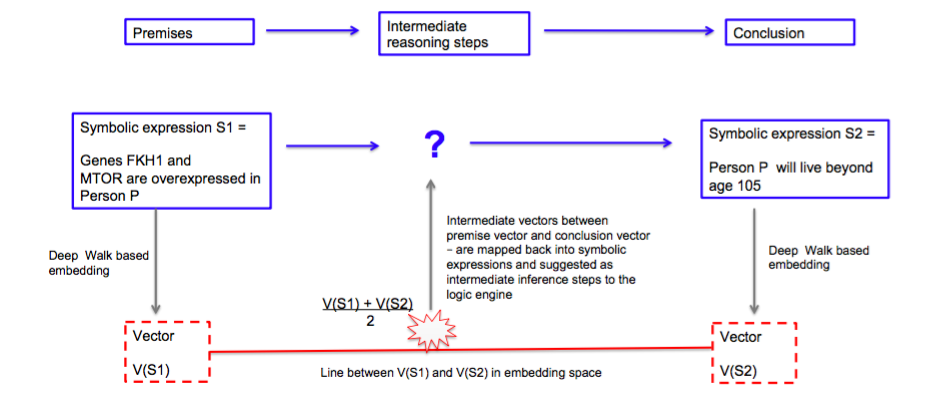}
\caption{Illustration of the concept of using differences in embedding vector space to guide the direction of uncertain intensional inference.}
\label{fig:inference}
\end{figure*}

Among the many potential applications of this correspondence between vector difference relations and intensional logic relations, is the use of vector algebra to guide inference control.   If one has premises and a hypothetical conclusion, and wants to explore inferences leading from the premises to the conclusion, it may be interesting to look at the vector pointing from the embedding vector of the premises to the embedding vector of the conclusion (i.e. the vector conclusion - premises).   Points along this vector may correspond to Atoms that are promising to consider as intermediary steps in inferences leading from the premises to the conclusion.   Figure \ref{fig:inference} illustrates this notion, which is a current focus of research.

\section{Conclusion and Future Work}

We have presented early exploratory work into potential close alignment between relationships among embedding vectors corresponding to nodes in a semantic hypergraph and uncertain intensional logic relationships among these nodes.   Next steps include systematically evaluating the prevalence and strength of these mappings, validating their generalization into a broader functorial mapping, exploring them in contexts beyond biology such as everyday commonsense reasoning and mathematical theorem-proving, and leveraging these relationships for guidance of inference control.

 \bibliographystyle{splncs04}
%




\bibliography{bbm.bib}
\end{document}